# RESEARCH OF THE ROBOT'S LEARNING EFFECTIVENESS IN THE CHANGING ENVIRONMENT

*V.Ya. Vilisov*
*Energy IT LLP, University of Technology, Russia, Moscow Region, Korolev*
*vvib@yandex.ru*

**Abstract.** *The object of the research is the adaptive algorithms that are used by the operator when educating the robotic systems. Operator, being the target-setting subject, is interested in the goal that robotic systems, being the conductor of his targets (criteria), would provide a maximum effectiveness of these targets' (criteria's) achievement. Thus, the adaptive algorithms provide the adequate reflection of the operator's goals, found in the robotic systems' actions. This work considers potential possibilities of such target adaption of the robotic systems used for the class of the allocation problems.*

### Introduction

Lately, the tendency of the robotic intellectualization becomes more and more evident. At that several directions are being developed: as the traditional bionic direction, implemented in the construction of animats [2], so and variants of learning with a teacher [3, 4]. One of the significant aspects of the robotic intellect increase is the ability of the robots to adjust (to adapt) to the different environmental factors. This work is devoted to one of the robotic adaptation directions, i.e. spotting of the non-evident target preferences of the teacher (operator or the decision-maker) while monitoring his decisions [1]. Still, it is necessary to note that the adaptation mechanism, together with other methods of the parameters' setting, require some time lag. It is clear that should the modification dynamics of the operator's targets and processes be rather high, the applied control model may be late in re-educating itself. Therefore, it would be only natural to suggest that there exists some limit of the target dynamics, and should it be crossed, the setting algorithm will not be already able to get adapted to it. Thus, this work considers the adaptation frontiers basing on the criteria (target) non-stationarity of the robotic systems' work.

### Task Definition

Allocation problems comprise a significant share of the problems that are solved by a separate robot or a group of robots [2, 3]. As a main context we shall take the problem of allocating the task or some limited resource within the robotic group. Problems of similar structure can also appear when managing a separate robot, for example a service cleaning robot or logistics robots that move the cargo along the road network or perform loading and loading works in the warehouse [3].

One of the LPP [5] formats can be of the following form. Objective function is:

$$L(\bar{x}) = \sum_{j=1}^{n} c_j x_j. \quad (1)$$

First choice selection criterion; assuming it is the operator who maximizes the objective function:

$$\bar{x}^* = arg \max_{x_j} L(\bar{x}), \quad (2)$$

where $\bar{c} = \|c_j\|_n$ is the objective function coefficient vector; $\bar{x} = \|x_j\|_n$ is the vector of variable quantities; $\bar{x}^*$ is the best solution.

LPP functions can be often presented by two groups of inequations - one group reflects the limits in relation to the distributed resources:

$$A\bar{x} \leq \bar{a}_0, \quad (3)$$

where $A = \|a_{ij}\|_{mn}$ is the resources demand matrix; $\bar{a}_0 = \|a_{i0}\|_m$ is the vector of the resources that are available for distribution. Another group represents limits in relation to the variation range of the variable quantities:

$$\bar{x} \leq \bar{b}, \quad \bar{x} \geq 0. \quad (4)$$

(1) − (4) correlations represent the choice selection model, where, due to the flexible character of a practically any operation, performed by the modern robotic systems, the prior (and current) uncertainty is concentrated in $\bar{c}$ vector. Every new situation of the solution selection (management) is defined by the exact values of $\bar{a}_0$ and $\bar{b}$ vectors, which, as a rule, can be measured, reflecting the environmental condition. Structure and internal characteristic of the robotic systems are reflected in $A$ matrix, which stays known and practically unchanged during the operation.

As during the performance of the operation, executed by the robotic systems, the distribution system of the limited resources selection is repeated manifold (with different limitations), the problem of the criteria dependence is expected to be solved by applying an adaptive LPP form [1], with the basis being the solution of the reverse LPP as well as the justification of $\bar{c}$ vector according to the realization results of chosen $\bar{x}$ solution. Evaluation of $\bar{c}$ vector that was received in the process of the reverse LPP solution is actually an approximation of the current LPP preferences, which can reflect a multitude of target indices, which, in their turn, are differently interconnected and a priori unpredictable. At that the settings (evaluation of $\bar{c}$ vector) are performed either via double-loop circuit, as a rule, in the *offline* mode, using the retrospective data of similar operations (or via active optimized experiment [1]), or in the *online* mode, if it is allowed by the technology and dynamics of the specific robotic system. Generally, the structure of the objective function that is approximating the LPP preferences can be a non-linear one; for example if it possesses all the characteristics of the classic utility function [5]. In this case the solution algorithms of the reverse problems can also consider such characteristics.

When solving reverse LPP, the following information can be used: (1) the one connected with the quality of the received and applied solution and (2) system and current data referring to the situation where the decision should made. In order to simplify the views, let us assume that the limitations (4) are unified and added to the limitations (3). In this case $\{A^k\}, \{\bar{a}_0^k\}, \{\bar{x}^k\}$ is the observation sequence (with $k$ being the number of observation or planning cycle) of the system and the data referring to the situation where the decision should made, while $\hat{c}$ is the current vector evaluating the coefficients of the integral objective function, which is approximating the LPP preferences.

**Setting Algorithm**

The task of constructing evaluations of the objective function according to the observations can be solved in different ways [1]. Let us analyze one of them, where the following function can be performed with every observation:

1. Using another situation $\bar{a}_0^k$ (situation where the decision should be made) the person who makes the decision chooses (by intuition or with some specific mechanisms) the $\bar{x}^k$ solution, which he considers the best in this situation (according to the multitude of its evident and non-evident indices).

2. The $\bar{x}^k$ solution is applied, with the results used for evaluation of the solution's quality according to the binary scale: good/bad.

3. The reverse LPP is then solved [1] (for "good" solutions), which causes the clarification of $\hat{c}$ vector that is responsible for evaluation of the objective function.

When there appears another situation that requires a solution, the cycle is repeated again from clause 1.

As it is shown in some of the author's works [1], the parameters setting of the preference model according to the observations can be performed rather quickly. Still, in some of the applications, the environment of the robotic systems can change rapidly. New environment can see a change of the value hierarchy in reference to the person, who makes the decision, and who, in his turn, is influencing the robotic system. New situation can also be caused by the changes happening in the robotic system itself, for example by the increase or decrease of its functionality. Such changes demonstrate non-stationarity of the situation (circumstances) where the robotic system is working. All the situational changes are caused by the fact that in the new circumstances, the person who is making the decision receives a new preference model (it can be presented as an objective function). Changes of the model parameters can go smoothly, but there also can be the cases when such changes possess a step-type nature. Any of these types of changes can be considered a demonstration of non-stationarity.

Should the robotic system start working in the stationary environment, its education up to the accepted level of the functioning effectiveness is taking some time τ. In cases of non-stationarity, the robotic system should be re-educated. It is necessary to note that during the re-education period, the effectiveness of its work will be far from the maximum level. At that, if the re-education period of the robotic system is defined by the education algorithm (controlled by the person who is making the decision concerning the robotic system, and taking different forms; for example, passive evaluation in the normal functioning mode or active probing of the environment and person, making the decision), the non-stationarity dynamics can be quite unpredictable. This is why, when planning the operations, it is very important to be aware of and to consider the robotic systems' frontiers in order to alleviate (via re-education) the negative impact of the objective non-stationarity.

In cases when the robotic system is acting within the conditions of the active counter force of another party, it is possible to set such a task of the specific non-stationarity creation, which would decrease its effectiveness as much as possible via counter reacting with the robotic system. Surely, the frontier evaluation depends significantly as on the specific type and structure of the robotic system, so and on the context of its usage and the problems it is required to solve. Therefore, considering the length limitations of this article, we shall use the simplest example to demonstrate the solution of the problem.

**Example**

Without providing a detailed description of the numerous imitational experiments, let us comment on the main elements, parameters and results. In the bi-dimensional LPP the coefficients of the true (imitating) objective function of the person making the decision, were represented by the vector $\bar{c} = [0.8 \quad 0.6]^T$ in the initial functioning position. Within the class of the problems under consideration, for the purposes of not losing the similarity, the vectors representing the parameters of the objective function, are applied in the standard form, i.e. possessing a unit length, with objective function not containing a constant component. If, after observing the decisions of the person who is making them, the robotic system evaluates vector $\bar{c}$, then the adaptation period $\tau$ is the value that comprises the observation steps up to the moment when the decisions made referring to the $\hat{c}$ evaluations with the set probability belief, shall coincide with the decisions, obtained from the imitating objective function. For the problems possessing different dimensionality of the $n$ variables space, this period will differ, so let us define it as $\tau_n$. Experimental (imitational) research allows building the dependence of the standard (i.e. equal to [0; 1] interval) average effectiveness of the decisions made by the robotic system as $L(t, n)$ function, where $t$ is a number of the observational step or the decision-making process. For some decisions of the space dimensionality, this function, built according to the results of the imitational experiment, is shown on Picture 1.

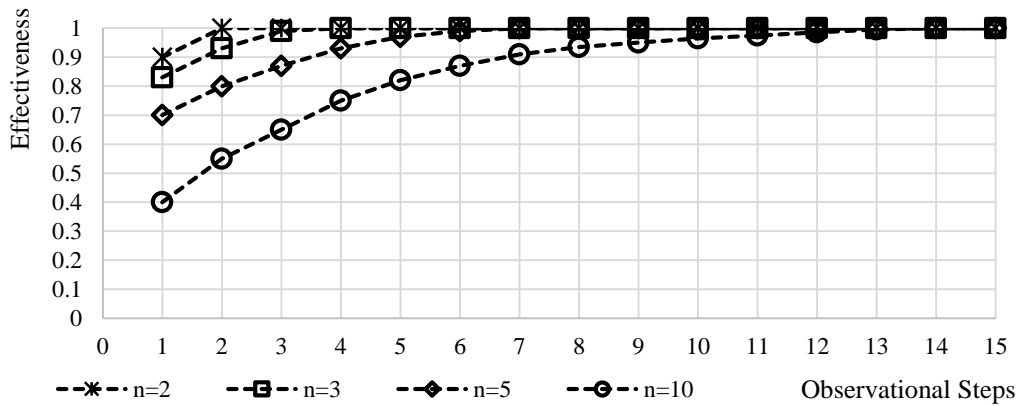

Picture 1. Setting dynamics of the models with different dimensionality

Non-stationarity was imitated by the step-type modifications of the vector that represented the parameters of the objective function. For example, the initial vector $\bar{c} = [0.8 \quad 0.6]^T$ rapidly transforms into the vector $\bar{c} = [0.6 \quad 0.8]^T$. During such a rapid change, the effectiveness also rapidly falls down to some level, going up together with the educational process of the model, used by the robotic system for solving the allocation problem. Thus, where there exist regular step-type change of the zero preferences (parameters of the objective function), the effectiveness chart of any dimensionality shall represent a saw-like line with the rapid falls of the effectiveness values and smooth increases, happening when the model of the robotic system is re-learning the preferences. In this situation, the effectiveness criterion of the robotic system's work can be presented by the average (in time) value of standard effectiveness.

**Conclusions**

1. Effectiveness of the robotic system's resource distribution models that are set from the experience of the person making the decisions, depends significantly as on numerous parameters of the functioning environment, so and on the adequacy level of the preference system that is provided by the person making the decisions regarding the robotic system and on whose behalf the robotic system performs different operations.

2. Impact of external and internal factors referring to the non-stationarity of target preferences can be alleviated by the re-education of the models that are used by the robotic systems for the controlling allocation problems. Still, there exist critical levels of the objectives' non-stationarity dynamics, which can lead to the significant reduction of the functionality of the robotic system's work.